\documentclass{article}

\usepackage{microtype}
\usepackage{booktabs} 

\usepackage{hyperref}
\usepackage{url}
\usepackage{amsthm}
\usepackage{amsfonts}
\usepackage[dvipsnames]{xcolor}
\usepackage{lipsum}  
\usepackage{multicol}
\usepackage{tikz}
\usetikzlibrary{bayesnet,shapes,arrows}

\usepackage{lipsum}
\usepackage{array}

\usepackage{amsmath}
\usepackage{amssymb}
\usepackage{mathrsfs}
\usepackage{graphicx}
\usepackage{subfig}
\usepackage{floatrow}
\usepackage{algorithm}
\usepackage{algorithmic}
\usepackage{float}

\newcommand{\bb}{\boldsymbol{b}}

\newcommand{\sbb}{\boldsymbol{s}}

\newcommand{\vb}{\boldsymbol{v}}

\newcommand{\xb}{\boldsymbol{x}}

\newcommand{\zb}{\boldsymbol{z}}

\newcommand{\zerob}{\boldsymbol{0}}
\newcommand\eqdef{\mathrel{\stackrel{\makebox[0pt]{\mbox{\normalfont\tiny def}}}{=}}}

\DeclareMathOperator*{\E}{\mathop{\mathbb{E}}}
\DeclareMathOperator*{\KL}{KL}

\usepackage[accepted]{icml2019}

\icmltitlerunning{Contrastive Variational Autoencoders}

\begin{document}

\tikzset{%
  every neuron/.style={
    circle,
    draw,
    minimum size=1cm
  },
  neuron missing/.style={
    draw=none, 
    scale=3,
    text height=0.333cm,
    execute at begin node=\color{black}$\vdots$
  },
}

\twocolumn[
\icmltitle{Contrastive Variational Autoencoder Enhances Salient Features}



\icmlsetsymbol{equal}{*}

\begin{icmlauthorlist}
\icmlauthor{Abubakar Abid}{stanfordee}
\icmlauthor{James Zou}{stanfordbds}
\end{icmlauthorlist}

\icmlaffiliation{stanfordee}{Department of Electrical Engineering, Stanford University, California, United States}
\icmlaffiliation{stanfordbds}{Department of Biomedical Data Science, Stanford University, California, United States}

\icmlcorrespondingauthor{James Zou}{jamesz@stanford.edu}

\icmlkeywords{Machine Learning, ICML}

\vskip 0.3in
]



\printAffiliationsAndNotice{
} 

\begin{abstract}

Variational autoencoders are powerful algorithms for identifying dominant latent structure in a single dataset. In many applications, however, we are interested in modeling latent structure and variation that are enriched in a target dataset compared to some background---e.g. enriched in patients compared to the general population.  Contrastive learning is a principled framework to capture such enriched variation between the target and background, but state-of-the-art contrastive methods are limited to linear models.
In this paper, we introduce the contrastive variational autoencoder (cVAE), which combines the benefits of contrastive learning with the power of deep generative models. The cVAE is designed to identify and enhance salient latent features.  The cVAE is trained on two related but unpaired datasets, one of which has minimal contribution from the salient latent features. The cVAE explicitly models latent features that are shared between the datasets, as well as those that are enriched in one dataset relative to the other, which allows the algorithm to isolate and enhance the salient latent features. The algorithm is straightforward to implement, has a similar run-time to the standard VAE, and is robust to noise and dataset purity. We conduct experiments across diverse types of data, including gene expression and facial images, showing that the cVAE effectively uncovers latent structure that is salient in a particular analysis. 
\end{abstract}

\section{Introduction}
In standard unsupervised learning, e.g. principal component analysis (PCA), the  goal is to identify subspaces or latent variables that capture the dominant variation in one dataset of interest. In many applications, however, we have multiple datasets and we are most interested in identifying patterns or variations that are \emph{enriched} in a target dataset compared to some background data. For example, in biomedical analysis, we are often interested in patterns that are specific to data from the treatment group that are not shared with the control group. Popular unsupervised learning methods work on each dataset separately and do not try to capture variation enriched in a particular dataset when contrasted against some background.  

Contrastive learning algorithms provide a way to identify such patterns. For example, contrastive principal components analysis (cPCA) is a recently-proposed method that seeks to identify \textit{salient} principal components in a target dataset by identifying those linear combinations of features that are enriched in that dataset relative to a background dataset, rather than those that simply have the most variance \citep{abid2018exploring}. The cPCA algorithm has been shown to discover latent factors in data that are missed by standard unsupervised learning methods including PCA, and has been extended to time-series decomposition \cite{dirie2018contrastive}. However, cPCA and its extensions have so far been limited to identifying linear patterns and subspaces within a dataset.

\begin{figure*}[!htb]
\centering
\subfloat[]{\includegraphics[width=0.32\linewidth]{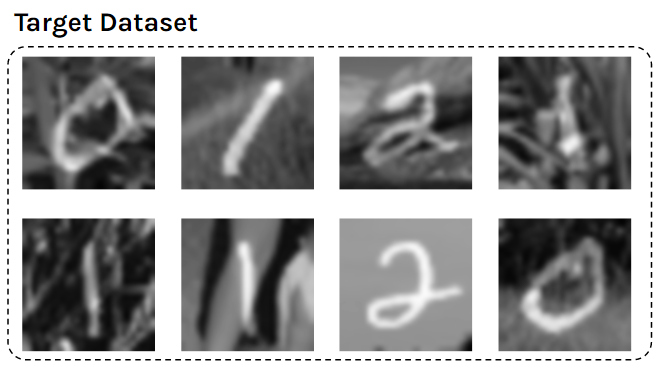}} \;
\subfloat[]{\includegraphics[width=0.32\linewidth]{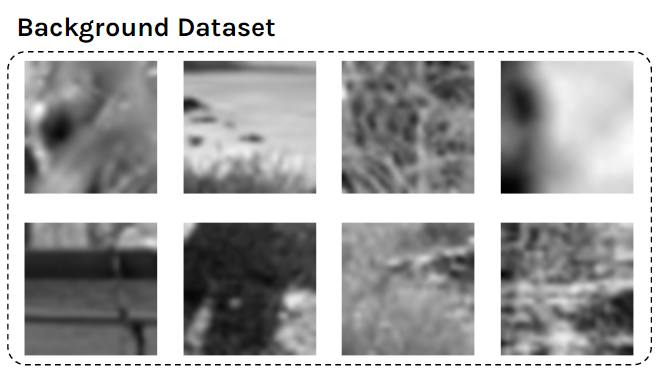}} \; 
\subfloat[]{\includegraphics[width=0.32\linewidth]{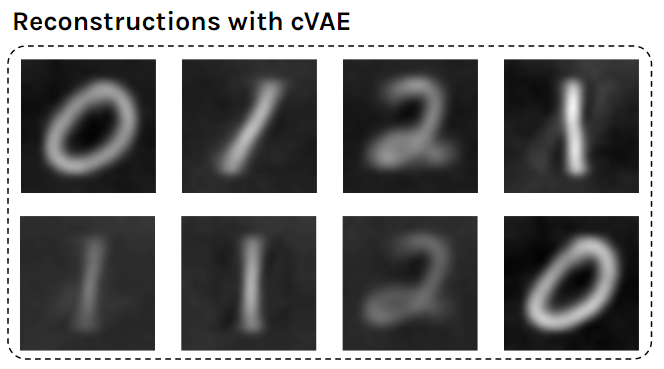}} 
\vspace{-0.3cm}
\caption{\textbf{Contrastive variational autoencoders applied to the Grassy-MNIST dataset.} (a) We construct a semi-synthetic dataset of hand-writen digits from MNIST dataset superimposed on images of grass from the ImageNet dataset (b) We assemble a separate background dataset of grass images, also from the ImageNet dataset, but not ones used in the creation of the target dataset. (c) We train a cVAE on the target and background datasets. The cVAE identifies that the salient latent variables are those related to the digits. We can remove the irrelevant latent variables to generate clean hand-written digits, despite the network never being trained on such images. See Appendix \ref{appendix:vae_results} for reconstructions with standard VAEs} \label{fig:mnist-examples}
\end{figure*}

In many complex datasets, such as those consisting of natural images or data collected from genomic assays, the key relationships between latent features and observed samples are highly nonlinear. Here, we develop the contrastive variational autoencoder (cVAE) to identify nonlinear latent variables that are enriched in one dataset relative to another.

\textbf{Motivating example: diverse dermatology images.} Consider a dataset of images of skin lesions that are collected from a diverse population in order to study disease-related variations in skin. Such a dataset will include images from individuals of different ethnicities and sexes. Training a standard variational autoencoder (VAE) and examining the dominant latent factors will likely identify features of the skin related to the demographic diversity of the subjects (e.g. skin color, and other features related to age and gender). More subtle features related to disease type and progression, which are of interest, may not appear as prominent latent factors, or may be entangled with the dominant latent factors. If we also have skin images from different, healthy participants, which contain image variations related to the demographics of the subjects but not relevant to diseases, we may be able to isolate these the salient disease latent factors by contrasting the disease and healthy image datasets. The cVAE algorithm that we develop in this paper offers a principled approach to do so. Note that the goal here is not to classify disease from healthy; such discriminatory learning approach often miss the complex variation within the disease group that we want to model \cite{abid2018exploring}. 

As a generative model, the cVAE can further be used to generate new samples with varying levels of the salient and irrelevant latent variables. For example, in Fig. \ref{fig:mnist-examples}, we show how a cVAE trained on images of hand-written digits on complex grassy backgrounds and as well as a background dataset of only grass images can be used to generate brand new images of clean hand-written digits, despite the cVAE never observing such images during training. We test cVAEs on several kinds of datasets, and find that, with a suitable background dataset, they consistently outperform standard VAEs in identifying and isolating latent features that are salient to the analyst. We have made the code for our experiments are publicly available\footnote{\url{https://github.com/abidlabs/contrastive_vae}}.

\subsection{Related Work}

Variational autoencoders (VAEs) are a powerful deep generative framework to capture latent structure in complex data \citep{kingma2013auto, doersch2016tutorial}.
Much recent work has focused on learning disentangled representations with VAEs \citep{kim2018disentangling, chen2018isolating}, in which each latent feature learns one semantically meaningful factor of variation, while being invariant to other factors. Other work has focused on learning relationships \textit{between} latent features, such as \citep{sonderby2016ladder}, which proposed learning latent features along with a dependency structure between them encoded as hierarchical network. 

In distinguishing between salient and irrelevant latent features, we adapt the principles of contrastive analysis that have been proposed in the context of unsupervised learning \cite{zou2013contrastive, ge2016rich}. These principles are utilized by the contrastive principal components analysis (cPCA) algorithm \citet{abid2018exploring}  to distinguish between salient and irrelevant principal components by introducing a background dataset without the salient variance. \citet{dirie2018contrastive} extended this idea to time series analysis by developing contrastive singular spectrum analysis (cSSA) to decompose a time series into salient subsignals that are absent from a secondary time series. Extending the ideas of contrastive analysis to VAEs was independently explored in concurrent work by \citep{severson2018unsupervised}, who has a more restricted formulation than the method proposed in this paper. 

Our work also bears some resemblance to recent work in domain adaptation, where the goal is to learn domain-specific representation separately from factors that are shared across domains \citep{bousmalis2016domain, gonzalez2018image}. However, these methods are not designed to extract salient latent features in one dataset in an unsupervised manner; the method proposed by \citep{bousmalis2016domain} is designed with intention of feeding shared features into a single classifier, while \citet{gonzalez2018image} primarily design their method, based on a pair of generative adversarial networks (GANs), as a means for sampling and image translation. Because the goals of these methods is different than ours, the architecture and loss functions are also considerably more complex, with a large number of components that are not needed for the cVAE algorithm.

\section{Problem Setting}

We begin by describing the generative latent variable model that motivates the development of the cVAE algorithm. Consider the samples $\{\xb_i\}_{i=1}^n$ in our target dataset to be $n$ i.i.d. samples that are generated by a random process that involves two sets of latent variables $\zb_i$, $\sbb_i$, which are drawn from a known prior distribution. The observed samples are drawn from a conditional distribution that is an arbitrary function $f_{\theta}$ of the latent variables. This function is parametrized by unknown parameters $\theta$:
\begin{align}
\xb_i \sim f_\theta(\xb \vert \zb_i, \sbb_i)
\end{align}
We use $\sbb$ to refer to the latent variables that are \textit{salient} in a particular analysis. In the example presented in the introduction, these may be the latent variables related to subtypes of skin cancer, or the progression of the disease. The second set of variables $\zb$ are the latent variables \textit{irrelevant} for the analysis, such as features related to demographic variations among the individuals. 

Trained on such a dataset, a standard VAE will not necessarily discover the salient latent features. In fact, if the irrelevant latent features generate most of the variation in the data, the VAE will identify those irrelevant features as they allow the VAE to minimize the reconstruction loss more effectively. Alternatively, the VAE may learn latent representations in which the irrelevant variables are entangled with the relevant latent variables.

\begin{figure}[!bt]
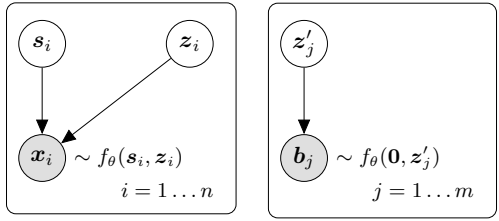
 
\centering
\vspace{0.5cm}
\resizebox{0.8\linewidth}{!}{
  \tikz{ %
        \tikzset{plate caption/.append style={below left=0pt and 0pt of #1.south east}}
        \node[latent] (s) {$\sbb_i$} ; %
        \node[latent, right= 1.5cm of s] (z) {$\zb_i$} ; %
        \node[obs, below=of s, label=0:{$\sim f_\theta(\sbb_i, \zb_i)$}] (x) {$\xb_i$} ; %
        \path[->]
        (s) edge node[right] {} (x) 
        (z) edge node[right] {} (x);
        \plate[inner sep=0.25cm, xshift=0.08cm, yshift=0cm] {plate1} {(x) (z)} {$i=1 \ldots n$}; %
        
        \node[latent, right= 1cm of z] (r) {$\zb_j'$} ; %
        \node[obs, below=of r, label=0:{$\sim f_\theta(\zerob, \zb_j') $}] (b) {$\bb_j$} ; %
        \coordinate[above right= 0.1cm and 2.3cm of r] (rs);
        \path[->]
        (r) edge node[left] {} (b); 
        \plate[inner sep=0.25cm, xshift=0.08cm, yshift=0cm] {plate2} {(rs) (b)} {$j=1 \ldots m$}; %
        
      }}
    \caption{\textbf{Generative model for target and background}. The generative model for each sample of the target data $\xb_i$ and background data $\bb_j$, includes two sets of latent variables, $\sbb, \zb \sim \mathcal{N}(0, I)$. The former consists of those latent variable that are salient to the analyst, and the latter consists of those variables which are irrelevant. Here, $\xb_i \sim f_\theta(\sbb_i, \zb_i)$. For the background dataset, the presence of the salient latent variables is minimal or absent, so we have  $\bb_j \sim f_\theta(\zerob, \zb_j')$. Observed quantities are shaded. \label{fig:contrastive_generative_model}}
\end{figure}

How can we identify and isolate the salient latent variables? Following the contrastive analysis framework, we introduce a secondary background dataset $\{\bb_j\}_{j=1}^m$ in which the salient latent features are absent:
\begin{align}
\bb_j \sim f_\theta(\bb \vert \zerob, \zb_j').
\end{align}
Let us assume that the unobserved random variables $\sbb_i, \zb_i, \zb_j$ are independently drawn from the isotropic Gaussian prior, and here, we use different indices $i$ and $j$ to indicate the the target and background datasets are \textit{unpaired}, though the samples in each dataset come from the same generative process $f_\theta(\cdot)$. The generative model is shown as a probabilistic graphical model in Fig. \ref{fig:contrastive_generative_model}.

Background datasets naturally exist in many settings. For example, we have observed that in datasets collected from diseased patients, where the salient latent factors are those related to disease type and progression, a control group of healthy subjects can provide a suitable background. More generally, when population-level variations are to be excluded, one can often use samples drawn from a similar population as the target dataset, but without the variation of interest. Conversely, a homogeneous sample can be used as the background for a sample drawn from a mixed population to eliminate the intra-group variation and measurement noise, while enhancing the inter-group variation. In other settings, such as the Grassy-MNIST dataset introduced in Fig. \ref{fig:mnist-examples}, a set of signal-free data can serve as a good background to remove structured noise from samples that include both signal and noise.

The learning problem that we are faced with then becomes designing an algorithm for learning the parameters $\theta$ of the transformation $f_\theta(\cdot)$ from unpaired target and background datasets. We are also interested in approximate posterior inference of the latent features. In particular, we seek to  distinguish the salient latent features $\sbb$ and the irrelevant latent features $\zb$.

\section{Method}
\label{section:method}

\begin{figure*}[!htb]
\centering
\subfloat[]{\raisebox{8mm}{\includegraphics[width=0.47\linewidth]{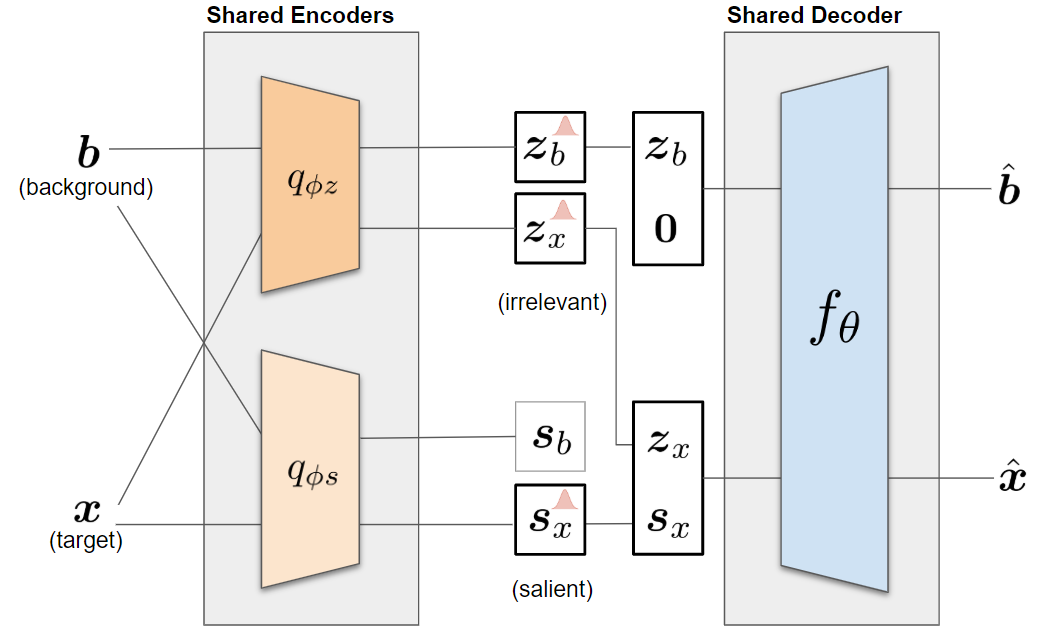} }} 
\subfloat[]{\includegraphics[width=0.5\linewidth]{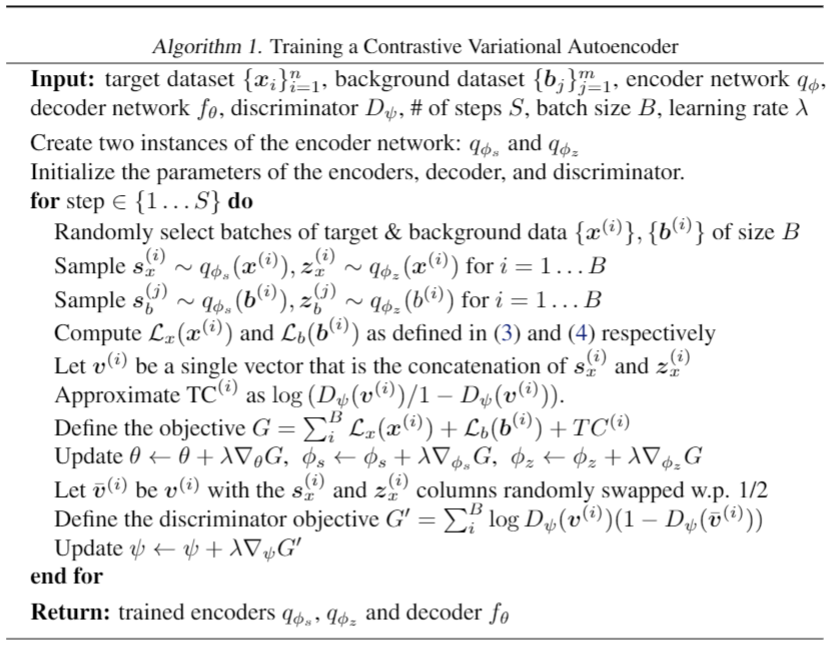}} 
\vspace{-0.3cm}
\caption{\textbf{Architecture and algorithm for the cVAE.} (a) Here, we illustrate the architecture of the contrastive VAE. The background target datasets are fed into two sets of shared encoders $q_{\phi_s}$ and $q_{\phi_z}$, which output the salient and irrelevant latent variables respectively. The latent variables for the target dataset are concatenated and fed into a decoder network, while only the (zero-padded) irrelevant variable is fed into the decoder network. The decoder outputs reconstructions of the target and background data. (b) The pseudocode for the cVAE is provided here,  using batch gradient descent for simplicity.} \label{fig:method}
\end{figure*}

To approach this problem, we introduce two probabilistic encoders, denoted by $q_{\phi_s}(\sbb \vert \xb)$ and $q_{\phi_z}(\zb \vert \xb)$,  which are parametrized approximations to the intractable posteriors over the latent variables $\sbb$ and $\zb$ respectively. We can derive a variational bound on the likelihood of individual data points, in terms of the unknown parameters, similar to that of the standard variational autoencoder \citep{kingma2013auto}. For the target dataset, we have the following likelihood lower bound (see Appendix \ref{appendix:derivation} for derivation):
\begin{align}
    \label{eq:target_likelihood}
    \mathcal{L}_x(\xb_i) &\ge \E_{q_{\phi_s}(s) q_{\phi_z}(z)}[f_\theta(\xb_i | \sbb, \zb)] -  \KL(q_{\phi_s}(\sbb | \xb_i) \Vert p(\sbb)) \nonumber \\ &- \KL(q_{\phi_z}(\zb | \xb_i) \Vert p(\zb)) 
\end{align}
For the likelihood of the background dataset, we have:
\begin{align}
    \label{eq:background_likelihood}
    \mathcal{L}_b(\bb_i) \ge \E_{q_{\phi_z}(z)}[f_\theta(\xb_i | \zerob, \zb)] - \KL(q_{\phi_z}(\zb | \bb_i) \Vert p(\zb)) 
\end{align}
Now, we train neural networks to maximize the sum of objective functions (\ref{eq:target_likelihood}) and (\ref{eq:background_likelihood}) by training two encoder networks $q_{\phi_s}$ and $q_{\phi_z}$, to infer $\sbb$ and $\zb$ respectively from the observed features, and a single decoder network $f_\theta(\cdot)$ which takes as input a concatenated version of $\xb$ and $\zb$ as input and reconstructs the original samples. Such an architecture is shown in Fig. \ref{fig:method}(a). This allows to use stochastic gradient descent on mini-batches of the target and background dataset to learn the parameters of the three neural networks.

Empirically, we observe that we can further improve results by encouraging independence between salient and irrelevant latent features. We add a \textit{total correlation} (TC) term to the objective function for the target dataset, 
\begin{align}
\label{eq:tc_term}
\text{TC} = - \KL(\bar{q} \Vert q_{\phi_s}(\sbb | \xb_i) \cdot q_{\phi_z}(\zb | \xb_i)),
\end{align}
where $\bar{q} \eqdef q_{\phi_s, \phi_z}(\sbb, \zb | \xb_i)$ is the joint conditional probability of $\sbb, \zb$. If $\sbb$ and $\zb$ are independent, then (\ref{eq:tc_term}) is zero; otherwise, it is negative. However, calculating these terms over mini-batches of samples is difficult to do directly. To approximate this term, we use the density-ratio trick, analogous to the method described in \citet{kim2018disentangling}. In essence, we use a separate discriminator to distinguish samples $[\sbb, \zb]$ drawn from $\bar{q}$ and from $q_{\phi_s}(\sbb | \xb_i) \cdot q_{\phi_z}(\zb | \xb_i)$. The classification probabilities of this discriminator can be used to estimate (\ref{eq:tc_term}). This discriminator is trained simultaneously with the encoder and decoder neural networks described earlier. More details are provided in Appendix \ref{appendix:total_correlation}.

Once the parameters $\phi_s$, $\phi_z$, and $\theta$ are learned, the cVAE can be used to solve three related problems that make it particularly useful for settings in which there are both salient and irrelevant latent features:
\begin{enumerate}
    \item \textbf{Infer only the salient features} by feeding a sample $\xb_i$ into $q_{\phi_s}$ and sampling from the resulting distribution. 
    \item \textbf{Generate samples using only the salient features} by sweeping values of $\sbb_i$, concatenating $\sbb_i$ with $\zerob$, and feeding the result into the decoder $f_\theta$.
    \item \textbf{Denoise samples (i.e. remove irrelevant structure)} by feeding a sample $\xb_i$ into $q_{\phi_s}$ and sampling $\sbb_i$ from the resulting distribution, then concatenating $\sbb_i$ with $\zerob$, and feeding the result through the decoder $f_\theta$.
\end{enumerate}

For the latter two tasks, notice that we have set $\zb_i = 0$ to simulate the effect of removing the irrelevant latent features. For linear encoders and decoders (i.e. PCA), this would remove any influence of the irrelevant latent variables. However, for neural networks, this is not generally the case, since a neural network may transform an input of $\zerob$ to a non-zero output. To ensure that this does not happen, in experiments where we denoise samples, we constrain all of the neural networks $q_{\phi_x}$,  $q_{\phi_z}$, and $f_\theta$ to have zero biases. Coupled with ReLU activations, this ensures that an input of $\zerob$ to any of these networks does not result in a  non-zero output. Pseudocode to train cVAEs is shown in Fig. \ref{fig:method}(b).

\begin{figure*}[!htb]
\centering
\subfloat[]{\includegraphics[width=0.21\linewidth]{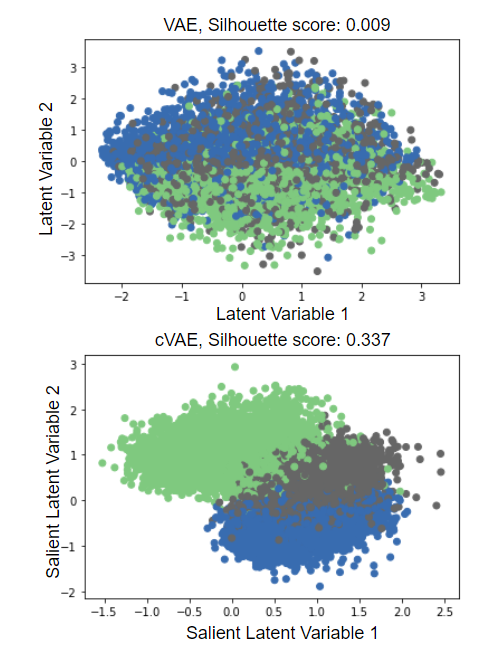}} \;
\subfloat[]{\includegraphics[width=0.175\linewidth]{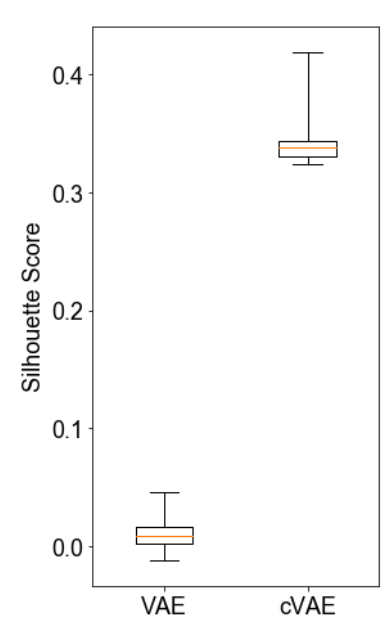}} \; 
\subfloat[]{\includegraphics[width=0.285\linewidth]{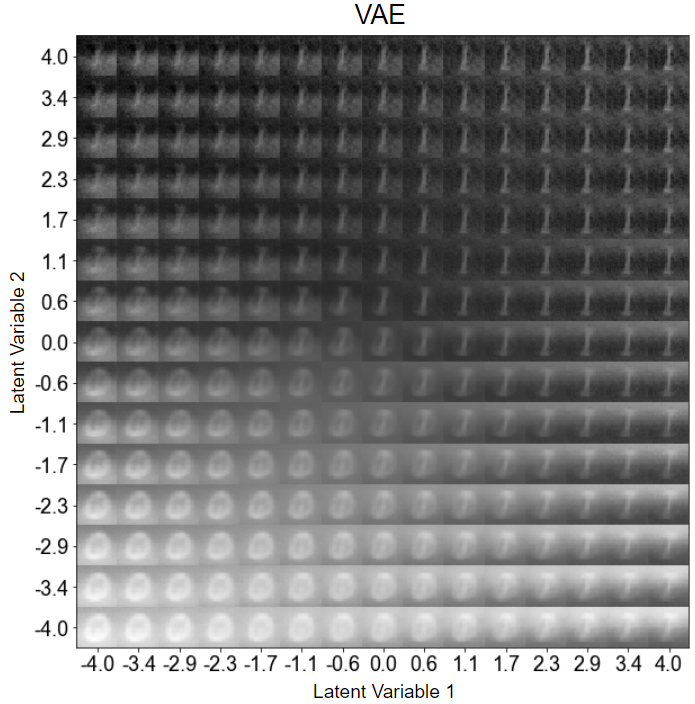}} \;
\subfloat[]{\includegraphics[width=0.28\linewidth]{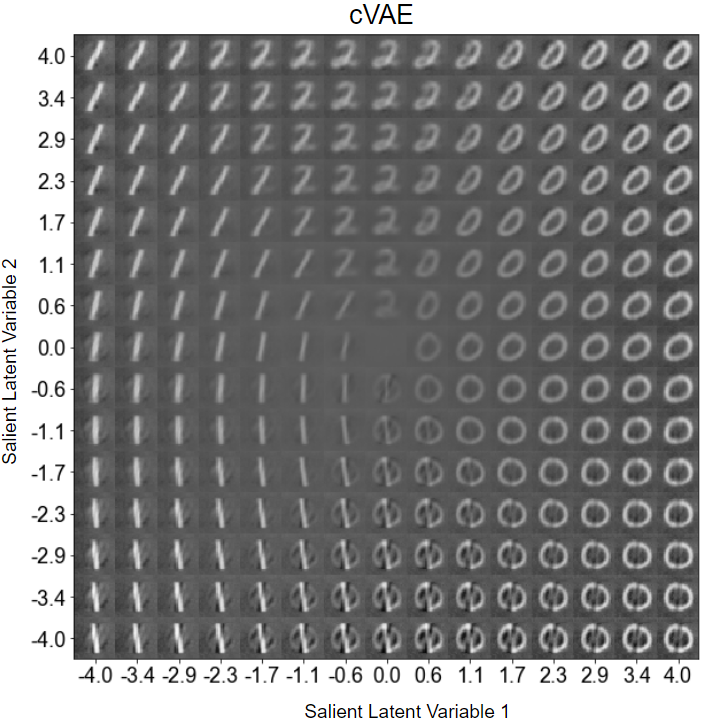}} 
\vspace{-0.3cm}
\caption{\textbf{Evaluation of the VAE and cVAE on the Grassy-MNIST dataset.} Here, we show the results of applying both the standard VAE and cVAE to the Grassy-MNIST dataset, introduced in Fig. \ref{fig:mnist-examples}. (a) The top panel shows that when the samples are embedded into the 2-dimensional latent space of the VAE, they do not cluster by digit (here, images with the digit 0 are in \textcolor{YellowGreen}{green}, with the digit 1 are in \textcolor{teal}{blue}, and the digit 2 in \textcolor{gray}{gray}). The bottom panel shows that when the samples are embedded into the 2-dimensional $s$-latent space of the VAE, they do cluster by digit. (b) These results are also  consistent, as shown in this boxplot which shows the resulting silhouette scores across 10 trials. (c) Here, we use the trained VAE to generate new samples by sweeping values in the 2-dimensional latent space. The generated samples include both digit features and background features. (d) Here, we use the trained cVAE to generate new samples by sweeping values in the 2-dimensional latent space resulting in clean hand-written digits. In Appendix \ref{appendix:high_dimension_vae}, we show that even a standard VAE with a larger latent space does not learn digit-related factors in this setting.} \label{fig:mnist-quant}
\end{figure*}

\section{Evaluation}
\label{section:evaluation}

We evaluate cVAEs on several target and background datasets, both quantitatively and qualitatively. On the quantitative side, we choose datasets in which there exist ground-truth classes, but which are not detected easily using only the dominant latent features. We train both standard VAEs and cVAEs and evaluate whether the samples, when projected into the learned latent space, produce distinct clusters in agreement with classes that present in the target dataset. We use this method to evaluate because it represents the unsupervised learning regime, in which an analyst without class labels may  visualize or embed points in a lower dimension using VAEs to reveal salient substructure.

We use the \textit{silhouette score} (SS)  to quantify the degree to which the clustering agrees with ground-truth class labels \citep{rousseeuw1987silhouettes}. This metric has been used in prior literature to assess the quality of latent VAE embeddings \citep{hu2018parameter, clachar2016novelty}. The SS is defined as follows: for a single data point $i$, let  $SS(i) \eqdef \frac{b(i) - a(i)}{\max\{a(i), b(i)\}}$, where $a(i)$ is the average distance from $i$ to the other points that share its label, while $b(i)$ is the average distance from $i$ to the points in the next nearest cluster. The SS for a dataset is the mean $SS(i)$ for all of the points $i$ in the dataset, and varies between $-1$ and $1$, with a higher positive value indicating that clusters agree more strongly with class labels. 

On a qualitative level, we have shown in Fig. \ref{fig:mnist-examples} that the cVAE is able to denoise images, removing complex grassy backgrounds present in a target dataset of hand-written digits. We further show that the cVAE can be used to generate samples in which only the salient features vary, making the cVAE useful as a way to precisely adjust salient features independently of irrelevant ones when generating samples. 

\subsection{Target and Background Datasets} 

\textbf{Grassy-MNIST.} We constructed this semi-synthetic target dataset by superimposing hand-written digits from the popular MNIST dataset \citep{lecun2010mnist} with images labeled as grass in the ImageNet \citep{deng2009imagenet} dataset. Each image of grass was cropped, resized to be 28 by 28 pixels, converted to grayscale, and scaled to have double the amplitude of the MNIST digits before being superimposed on the digit. Only the digits 0, 1, and 2 were used in the target dataset. The background dataset consisted of other images of grass not used in the construction of the target dataset.

\textbf{RNA-Seq with Batch Effects.} We combined two single-cell RNA-Seq datasets released by \citet{zheng2017massively} measuring the expression levels of genes from blood cells collected from a leukemia patient before and after transplant. The samples were labeled by which step (before/after the procedure) the cells were collected. These cells contained significant batch effects, so as a background, we selected a dataset of healthy blood cells released by the same authors and processed on the same equipment.

\textbf{CelebA. } We used the popular celebrity faces dataset \citep{liu2015faceattributes} after cropping each image and scaling it to be 64 by 64 pixels. We used subsets of the CelebA dataset with certain attributes as the target. Namely, we formed one target dataset of celebrities wearing glasses, and we formed a second target dataset of celebrities wearing caps. In both cases, the remaining celebrities constituted the background. 

\subsection{Implementation Details}

The VAE and cVAE models were implemented in Tensorflow\footnote{\url{https://www.tensorflow.org/}}, using the Keras API. The target and background datasets were fed simultaneously into the model. The loss function consisted of the sum of the negative of the terms in  (\ref{eq:target_likelihood}) and (\ref{eq:background_likelihood}), along with the TC loss, and the loss of a discriminator used to estimate the TC (see Appendix \ref{appendix:total_correlation} for more details). We used logistic regression as the discriminator.

We used the Adam optimizer \citep{kingma2014adam} with a learning rate of $0.001$, $\beta_1 = 0.9$, $\beta_2 = 0.999$. We used a batch size of 128 for our experiments. For the Grassy-MNIST and RNA Seq datasets, the VAE architecture consisted of fully-connected layers in the encoder and decoder. For the CelebA dataset, a convolutional VAE architecture was used. The exact architecture of our models can be found in Appendix \ref{appendix:architecture}. 

\subsection{Quantitative Results}

We first trained a standard VAE on the Grassy-MNIST target dataset with a latent dimensionality of 2. We embedded each sample in the latent space, producing the plot shown in Fig. \ref{fig:mnist-quant}(a, top), in which each sample is colored by the digit that appears in the image. In this latent space, the samples do not cluster by the digit (SS=0.009); this is because the dominant features are those that relate to the grassy background, rather than features of the hand-drawn digit.

We then trained a cVAE on the Grassy-MNIST target and background datasets with a $s$-latent dimension of 2 and an $z$-latent dimension of 2. We embedded each point in the dataset onto the 2D $s$-latent space, producing the plot shown in Fig. \ref{fig:mnist-quant}(a, bottom). We find that in this latent space, samples clustered strongly by digits (SS=0.337), indicating that the cVAE has identified digit-related features as the salient latent variables. This pattern is consistently reproducible; the results of 10 trials are shown in Fig. \ref{fig:mnist-quant}(b). 

\begin{figure}[!htb]
\centering
\subfloat[]{\includegraphics[width=0.5\linewidth]{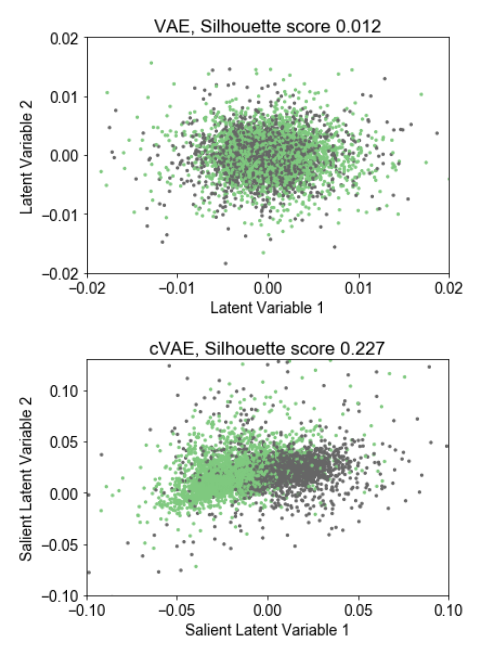}} \;
\subfloat[]{\includegraphics[width=0.422\linewidth]{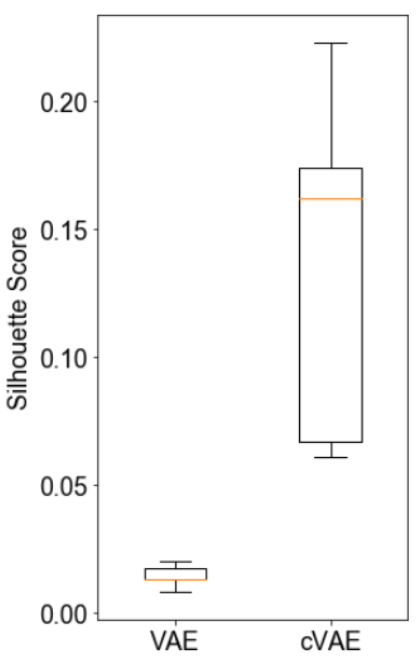}} 
\vspace{-0.3cm}
\caption{\textbf{Comparing VAE and cVAE on the RNA-Seq dataset.} Here, we show the results of applying both the standard VAE and cVAE to the RNA-Seq dataset. (a) The top panel shows that when the samples are embedded into the 2-dimensional latent space of the VAE, they do not cluster by transplant (here, pre-transplant samples are in green and post-transplant samples in gray). The bottom panel shows that when the samples are embedded into the 2-dimensional $s$-latent space of the cVAE, they do cluster by transplant status. A few points are omitted for ease of visualization; see complete plots and further analysis in Appendix \ref{appendix:fig_5}. (b) Aggregated silhouette scores over 10 trials are shown here.} \label{fig:rna}
\end{figure}

Next, we trained a standard VAE on the RNA-Seq dataset with a latent dimensionality of 2. We embedded each gene expression vector in the latent space, producing the plot shown in Fig. \ref{fig:rna}(a, top), in which each sample is colored to indicate whether it was pre/post-transplant. In this latent space, the samples do not cluster by the digit (SS=0.012); this is likely because the dominant features are those that relate to the batch effects or variations within cell types.

We then trained a cVAE on the RNA-Seq target and background datasets with a $s$-latent dimension of 2 and an $z$-latent dimension of 4. We embedded each point in the dataset onto the 2D $s$-latent space, producing the plot shown in Fig. \ref{fig:rna}(a, bottom). We find that in this latent space, samples clustered more strongly by pre/post-transplant status (SS=0.227), indicating that the cVAE was able to better identify latent features corresponding to transplant status. Aggregate results over 10 trials are shown in Fig. \ref{fig:rna}(b). 

\subsection{Qualitative Results}

\begin{figure*}[!htb]
\centering
\subfloat[]{\includegraphics[width=0.37\linewidth]{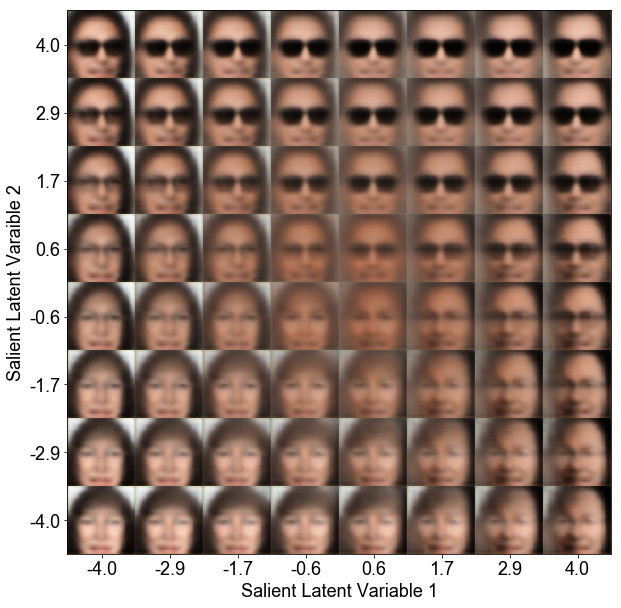}} \;
\subfloat[]{\includegraphics[width=0.115\linewidth]{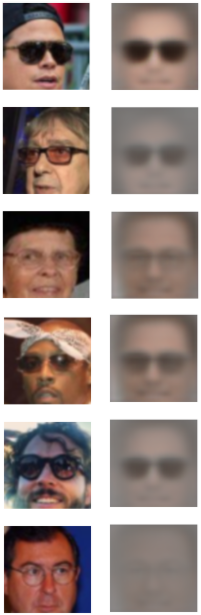}} \;
\subfloat[]{\includegraphics[width=0.37\linewidth]{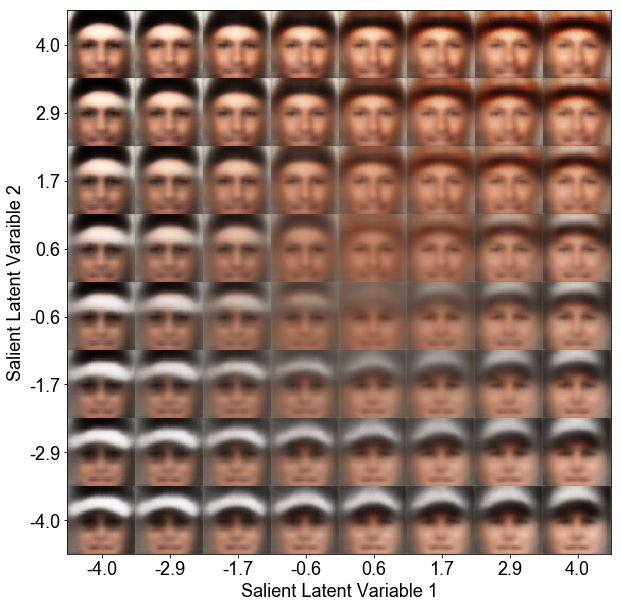}} \;
\subfloat[]{\includegraphics[width=0.11\linewidth]{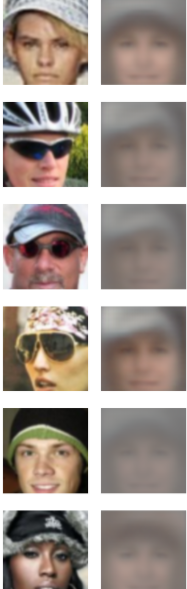}} 
\vspace{-0.3cm}
\caption{\textbf{Results of applying a cVAE to the CelebA Dataset.} Here, we apply cVAEs to a target dataset consisting of celebrity images either wearing glasses (for panels a-b) or caps (for panels c-d). In both cases, the remaining celebrities constituted the background dataset. (a) We use the trained cVAE to   generate new samples of celebrity images by varying the two learned latent factors. We see that the two latent factors concern attributes of the glasses. In particular, latent variable 1 seems to deal with glasses direction, while latent variable 2 is related to the transparency/presence of the glasses. (b) If we denoise the images by only selecting the salient features learned by the cVAE, we see that primarily the pixels that remain in the reoncstructed image relate to the glasses. (c) In this example, the cVAE has learned that the salient latent features that relate to attributes of the cap, namely the color of the cap and its position. (d)  If select only the salient features learned by the cVAE, we see that the pixels that relate to the cap in the original image are primarily present in the reconstruction, although some facial features are present as well. (\textit{cf.} figures in Appendix \ref{appendix:vae_results}, which show the results using standard VAEs.)} \label{fig:celeb}
\end{figure*}

Using the trained models on the Grassy-MNIST dataset, we generated new digits using the method described in Section \ref{section:method}. We show the results using the standard VAE and cVAE in Fig. \ref{fig:mnist-quant}(c) and Fig. \ref{fig:mnist-quant}(d) respectively. We find that the standard VAE generates new images with some degree of background structure. This background structure is entangled with digit features, with the digit `0' containing more background than the digit `1'. Furthermore, the digit `2` is not generated at all, as the features needed to reconstruct the digit are not learned by the standard VAE. On the other hand, we notice by using a contrastive VAE, we are able to generate the clean digits `0', `1', and `2', despite the network never being trained on such images. 

We then determine whether we are able to observe similar results on a more complex dataset, the CelebA dataset. In this case, we use two attributes, the presence of eyeglasses and the presence cap, to define our two target datasets. We use the remaining images as the background datasets. The purpose is to extract latent features related to the target attribute. For example, if we are interested in extracting features related to eyeglasses shape or color or size, we would define a target dataset of images of celebrities with glasses, and a background dataset in which none of the celebrities wear glasses.

We carry out such an analysis in Fig. \ref{fig:celeb}. In Fig. \ref{fig:celeb}(a), we find that the cVAE extracts salient features related to the eyeglasses, and in Fig. \ref{fig:celeb}(c), the cVAE extracts features related to the caps worn by the celebrities. Compared to results using standard VAEs (see Appendix \ref{appendix:vae_results}), we find that the cVAE more clearly extracts features that are related to the target attribute, and these are less entangled with dominant latent variables in the image, such as skin color and background color.  This is further supported by experiments shown in Fig. \ref{fig:celeb}(b) and Fig. \ref{fig:celeb}(d), in which we use the trained cVAEs to ``denoise'' the images, removing the effects of the irrelevant latent variables. We find, particularly in Fig. \ref{fig:celeb}(b), that this erases most of the pixels in the image, leaving only the features needed to reconstruct to the target attribute e.g. the glasses or the cap.

\subsection{Sensitivity Analyses}
\label{subsection:sensitivity}

In order to understand the reliability of cVAEs in diverse settings, we measured the robustness of cVAEs to various factors, such as the construction of the target/background datasets as well as the dimensionality of the latent space. We carried out experiments using the semi-synthetic Grassy-MNIST dataset. First, we varied the scale of the background images of grass used to construct the target images. We started with a scale of 0, representing no grassy images in the target data (i.e. the target was the original MNIST dataset) and, in increments of 0.25, increased the scale to 2 (i.e. the amplitude of the grassy background was twice that of the digits). The background dataset was unchanged.

\begin{figure*}[!htb]
\centering
\subfloat[]{\includegraphics[width=0.315\linewidth]{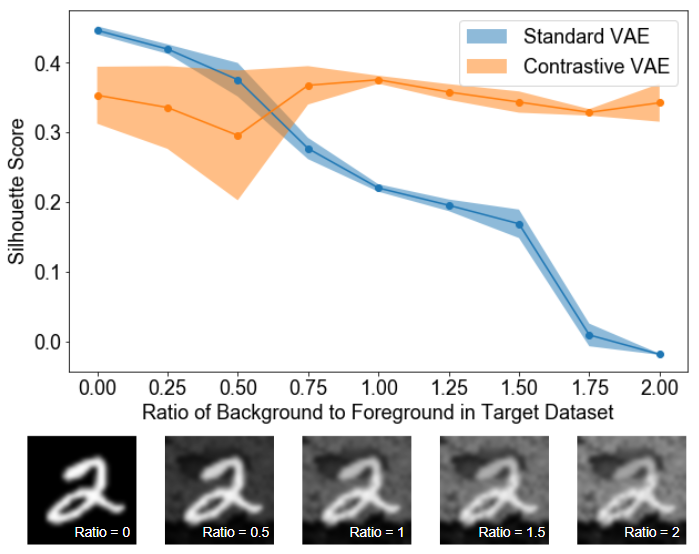}} \;
\subfloat[]{\includegraphics[width=0.32\linewidth]{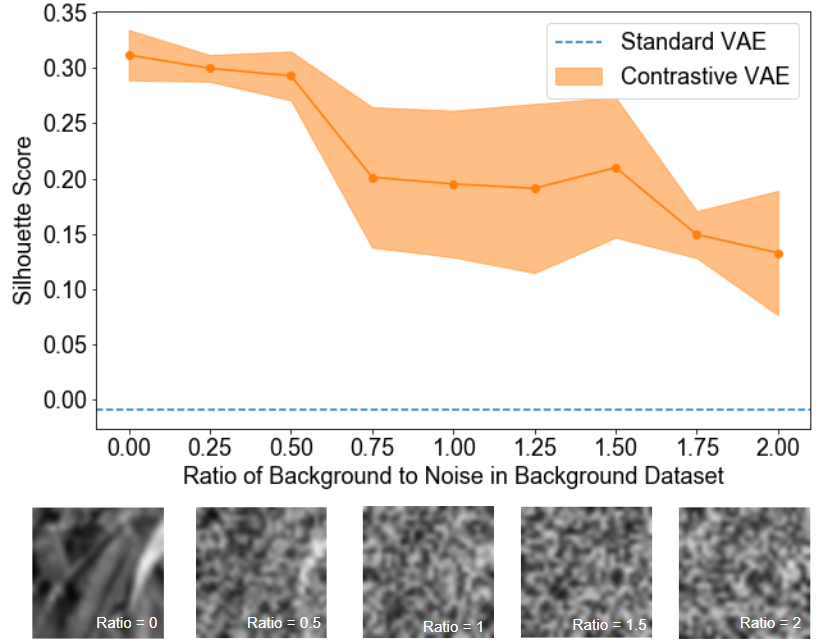}} \;
\subfloat[]{\includegraphics[width=0.24\linewidth]{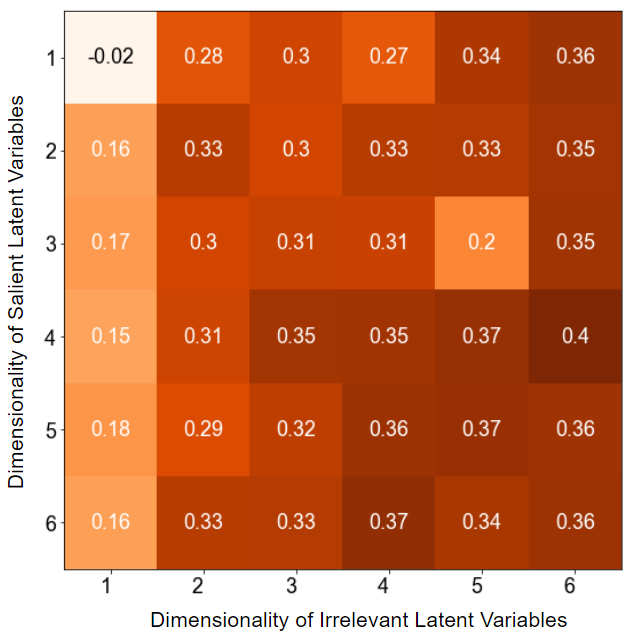}} 
\vspace{-0.3cm}
\caption{\textbf{Sensitivity of the cVAE to Dataset Composition and Hyperparameters.} Here, we probe the sensitivity of the cVAE algorithm to various factors. (a) We vary the construction the Grassy-MNIST dataset by change the relative amounts of `background' (images of grass), compared to the `foreground' (images of the digits) in each target image. We find that when there is little-to-no background, both kinds of VAEs perform well, with the standard VAE performing slightly better. With the introduction of the background, the performance of the standard VAE degrades, but not that of the contrastive VAE. Sample target images are shown below the plot. (b) Next, we add varying levels of isotropic noise to the background dataset, to see whether an imperfect background dataset affects performance. We do not observe significant reduction in performance even with the ratio of the aplitude of the noise-to-background increases. (c). Here, we explore the effect of changing the dimensionality of the $s$ and $z$ latent spaces. The Silhouette score for each combination of hyperparameters is listed in each spot, and the color scale reflects the amplitude of these scores. Further sensitivity analyses are included in Appendix \ref{appendix:sensitivity}.} \label{fig:sensitivity}
\end{figure*}

At each increment, we compared the efficacy of a standard VAE to that of a cVAE in identifying salient latent features. The results are shown in Fig. \ref{fig:sensitivity}(a). At low levels of background, both the standard VAE and the cVAE were effective at clustering MNIST digits. However, as the grassy background increased in amplitude, the efficacy of the standard VAE dropped significantly, while the cVAE continued to be able to discern relevant salient features.

We then considered how well-tailored a background dataset needed to be to serve as an effective contrast for a target dataset. We left the target dataset unchanged, but added varying levels of random isotropic noise to the background images of grass. We started with a noise level of 0,  representing no noise in the background data (i.e. the background images were the original images of grass from the ImageNet dataset), and in increments of 0.25, increased the scale to 2, at which point the images of grass were completely obscured by noise. Nevertheless, the performance of the cVAE did not seem to suffer significantly, as shown in Fig. \ref{fig:sensitivity}(b).

We also considered the importance of a hyperparameter in the cVAE algorithm: the dimensionality of the latent space used for $s$ and $z$. We varied each of these from 1-6 and trained a cVAE with those dimensions on the Grassy-MNIST dataset. We recorded the resulting silhouette scores in Fig. \ref{fig:sensitivity}(c). We found that increasing the size of the $z$-latent space consistently tended to improve results. Increasing the size of the $s$-latent space did not seem to improve or impair results. Additionally, we performed experiments to characterize the sensitivity of the cVAE to dataset composition; these experiments are described in Appendix \ref{appendix:sensitivity}.

\section{Discussion}
\label{section:discussion}

In this paper, we have introduced the cVAE, an extension of the standard VAE algorithm that allows users to identify latent factors that are salient to their own analyses through the use of a background dataset. We discuss the methodology behind the cVAE in Section \ref{section:method}, where we derive it using a variational approximation to the maximum likelihood problem. In Section \ref{section:evaluation}, we empirically compare the cVAE architecture to that of the standard VAE architecture on biological and image datasets, finding that the cVAE algorithm allows the user to: (1) identify salient latent factors, (2) generate new samples with only salient features, and (3) denoise samples by removing the irrelevant latent variables. 

We have shown, for example, that the cVAE can be used to discover latent factors that allow digits to be classified even in the presence of complex noise (see Fig. \ref{fig:mnist-quant}) or cells to be categorized even with strong batch effects (Fig. \ref{fig:rna}). In a dataset of celebrity images, cVAEs were not completely able to remove the effect of irrelevant latent variables, perhaps because they may be correlated with the salient latent variables in complex ways. However, they offered a way to enhance and more finely control the latent variables of interest (Fig. \ref{fig:celeb}).

It is worth emphasizing that the cVAE architecture is not designed for \textit{discriminative} analyses: the identified latent factors are not intended to discriminate betweeen points in the target and background dataset. Rather, the cVAE architecture helps the user identify and isolate salient latent factors that may be otherwise not present amongst or entangled with the dominant latent factors in a target dataset. 

The choice of the background dataset has a large influence on the result of cPCA. In general, the background data should have the structure or latent features that one would like to remove from the target data. Nevertheless, through systematic experiments, we find that the cVAE algorithm is fairly robust to noise in the background datasets, suggesting that even when the background dataset contains random latent factors that are not present in the target dataset, performance is not significantly adversely affected.

The primary strengths of the contrastive VAE is its generality and ease of use. Training a cVAE essentially requires the same runtime as a standard VAE and can be done in most situtations where VAEs are commonly used today. Doing so allows users to shift the focus from finding those latent features that are dominant in data to those that are salient in the context of their analyses.

\clearpage

\bibliography{main}

\begin{thebibliography}{21}
\providecommand{\natexlab}[1]{#1}
\providecommand{\url}[1]{\texttt{#1}}
\expandafter\ifx\csname urlstyle\endcsname\relax
  \providecommand{\doi}[1]{doi: #1}\else
  \providecommand{\doi}{doi: \begingroup \urlstyle{rm}\Url}\fi

\bibitem[Abid et~al.(2018)Abid, Zhang, Bagaria, and Zou]{abid2018exploring}
Abid, A., Zhang, M.~J., Bagaria, V.~K., and Zou, J.
\newblock Exploring patterns enriched in a dataset with contrastive principal
  component analysis.
\newblock \emph{Nature communications}, 9\penalty0 (1):\penalty0 2134, 2018.

\bibitem[Bousmalis et~al.(2016)Bousmalis, Trigeorgis, Silberman, Krishnan, and
  Erhan]{bousmalis2016domain}
Bousmalis, K., Trigeorgis, G., Silberman, N., Krishnan, D., and Erhan, D.
\newblock Domain separation networks.
\newblock In \emph{Advances in Neural Information Processing Systems}, pp.\
  343--351, 2016.

\bibitem[Chen et~al.(2018)Chen, Li, Grosse, and Duvenaud]{chen2018isolating}
Chen, T.~Q., Li, X., Grosse, R., and Duvenaud, D.
\newblock Isolating sources of disentanglement in variational autoencoders.
\newblock \emph{arXiv preprint arXiv:1802.04942}, 2018.

\bibitem[Clachar(2016)]{clachar2016novelty}
Clachar, S.
\newblock \emph{Novelty detection and cluster analysis in time series data
  using variational autoencoder feature maps}.
\newblock The University of North Dakota, 2016.

\bibitem[Deng et~al.(2009)Deng, Dong, Socher, Li, Li, and
  Fei-Fei]{deng2009imagenet}
Deng, J., Dong, W., Socher, R., Li, L.-J., Li, K., and Fei-Fei, L.
\newblock Imagenet: A large-scale hierarchical image database.
\newblock In \emph{Computer Vision and Pattern Recognition, 2009. CVPR 2009.
  IEEE Conference on}, pp.\  248--255. Ieee, 2009.

\bibitem[Dirie et~al.(2018)Dirie, Abid, and Zou]{dirie2018contrastive}
Dirie, A.-H., Abid, A., and Zou, J.
\newblock Contrastive multivariate singular spectrum analysis.
\newblock \emph{arXiv preprint arXiv:1810.13317}, 2018.

\bibitem[Doersch(2016)]{doersch2016tutorial}
Doersch, C.
\newblock Tutorial on variational autoencoders.
\newblock \emph{arXiv preprint arXiv:1606.05908}, 2016.

\bibitem[Ge \& Zou(2016)Ge and Zou]{ge2016rich}
Ge, R. and Zou, J.
\newblock Rich component analysis.
\newblock In \emph{ICML}, pp.\  1502--1510, 2016.

\bibitem[Gonzalez-Garcia et~al.(2018)Gonzalez-Garcia, van~de Weijer, and
  Bengio]{gonzalez2018image}
Gonzalez-Garcia, A., van~de Weijer, J., and Bengio, Y.
\newblock Image-to-image translation for cross-domain disentanglement.
\newblock \emph{arXiv preprint arXiv:1805.09730}, 2018.

\bibitem[Hu \& Greene(2018)Hu and Greene]{hu2018parameter}
Hu, Q. and Greene, C.~S.
\newblock Parameter tuning is a key part of dimensionality reduction via deep
  variational autoencoders for single cell rna transcriptomics.
\newblock \emph{bioRxiv}, pp.\  385534, 2018.

\bibitem[Kim \& Mnih(2018)Kim and Mnih]{kim2018disentangling}
Kim, H. and Mnih, A.
\newblock Disentangling by factorising.
\newblock \emph{arXiv preprint arXiv:1802.05983}, 2018.

\bibitem[Kingma \& Ba(2014)Kingma and Ba]{kingma2014adam}
Kingma, D.~P. and Ba, J.
\newblock Adam: A method for stochastic optimization.
\newblock \emph{arXiv preprint arXiv:1412.6980}, 2014.

\bibitem[Kingma \& Welling(2013)Kingma and Welling]{kingma2013auto}
Kingma, D.~P. and Welling, M.
\newblock Auto-encoding variational bayes.
\newblock \emph{arXiv preprint arXiv:1312.6114}, 2013.

\bibitem[LeCun et~al.(2010)LeCun, Cortes, and Burges]{lecun2010mnist}
LeCun, Y., Cortes, C., and Burges, C.
\newblock Mnist handwritten digit database.
\newblock \emph{AT\&T Labs [Online]. Available: http://yann. lecun.
  com/exdb/mnist}, 2, 2010.

\bibitem[Liu et~al.(2015)Liu, Luo, Wang, and Tang]{liu2015faceattributes}
Liu, Z., Luo, P., Wang, X., and Tang, X.
\newblock Deep learning face attributes in the wild.
\newblock In \emph{Proceedings of International Conference on Computer Vision
  (ICCV)}, 2015.

\bibitem[Nguyen et~al.(2010)Nguyen, Wainwright, and
  Jordan]{nguyen2010estimating}
Nguyen, X., Wainwright, M.~J., and Jordan, M.~I.
\newblock Estimating divergence functionals and the likelihood ratio by convex
  risk minimization.
\newblock \emph{IEEE Transactions on Information Theory}, 56\penalty0
  (11):\penalty0 5847--5861, 2010.

\bibitem[Rousseeuw(1987)]{rousseeuw1987silhouettes}
Rousseeuw, P.~J.
\newblock Silhouettes: a graphical aid to the interpretation and validation of
  cluster analysis.
\newblock \emph{Journal of computational and applied mathematics}, 20:\penalty0
  53--65, 1987.

\bibitem[Severson et~al.(2018)Severson, Ghosh, and
  Ng]{severson2018unsupervised}
Severson, K., Ghosh, S., and Ng, K.
\newblock Unsupervised learning with contrastive latent variable models.
\newblock \emph{arXiv preprint arXiv:1811.06094}, 2018.

\bibitem[S{\o}nderby et~al.(2016)S{\o}nderby, Raiko, Maal{\o}e, S{\o}nderby,
  and Winther]{sonderby2016ladder}
S{\o}nderby, C.~K., Raiko, T., Maal{\o}e, L., S{\o}nderby, S.~K., and Winther,
  O.
\newblock Ladder variational autoencoders.
\newblock In \emph{Advances in neural information processing systems}, pp.\
  3738--3746, 2016.

\bibitem[Zheng et~al.(2017)Zheng, Terry, Belgrader, Ryvkin, Bent, Wilson,
  Ziraldo, Wheeler, McDermott, Zhu, et~al.]{zheng2017massively}
Zheng, G.~X., Terry, J.~M., Belgrader, P., Ryvkin, P., Bent, Z.~W., Wilson, R.,
  Ziraldo, S.~B., Wheeler, T.~D., McDermott, G.~P., Zhu, J., et~al.
\newblock Massively parallel digital transcriptional profiling of single cells.
\newblock \emph{Nature communications}, 8:\penalty0 14049, 2017.

\bibitem[Zou et~al.(2013)Zou, Hsu, Parkes, and Adams]{zou2013contrastive}
Zou, J.~Y., Hsu, D.~J., Parkes, D.~C., and Adams, R.~P.
\newblock Contrastive learning using spectral methods.
\newblock In \emph{Advances in Neural Information Processing Systems}, pp.\
  2238--2246, 2013.

\end{thebibliography}
\bibliographystyle{icml2019}

\appendix
\newpage
\begin{onecolumn}

\section*{Appendices}

\section{Qualitative Results with Standard VAEs}
\label{appendix:vae_results}

In Fig. \ref{fig:mnist-examples}, we showed how cVAE reconstructions from the Grassy-MNIST dataset could be used to generate new clean images of hand-written digits never seen by the generative model. For comparison, here we show the reconstructions of a standard VAE:

\begin{figure}[H]
\centering
\includegraphics[width=0.39\linewidth]{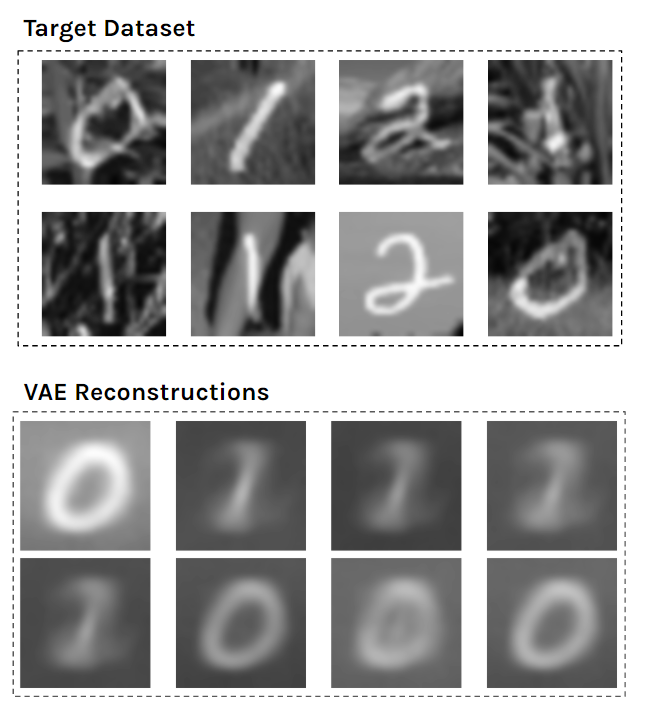}
\end{figure}

Next, we show the result of training a standard convolutional variational autoencoder with latent dimensionality of 2 on the CelebA dataset. Sweeping the latent variables produces the following images, which suggest that the dominant latent structure in the Celeb A dataset are related to skin color and background behind, or orientation of, the face.

\begin{figure}[H]
\centering
\includegraphics[width=0.39\linewidth]{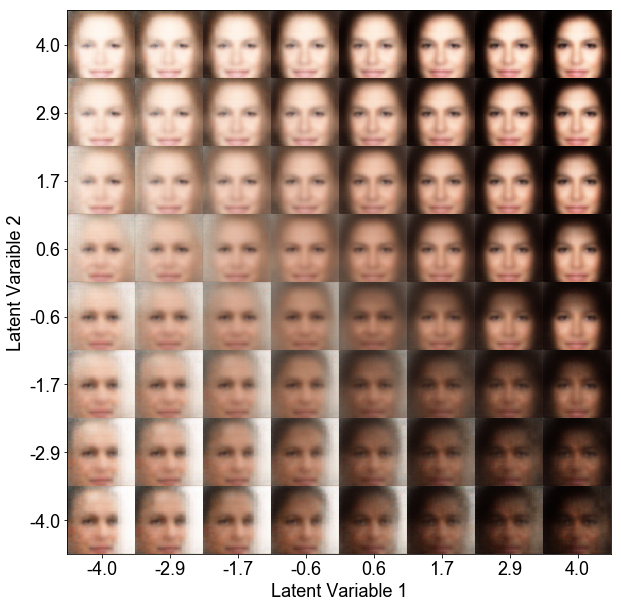}
\end{figure}

\clearpage

In Fig. \ref{fig:celeb}(a), we showed that we could isolate latent variables related to eyeglasses by using a contrastive VAE. For comparison, here we show the reconstructions of a standard VAE trained \textit{only} on the target dataset of celebrities with glasses. Because we are restricting ourselves to celebrities with glasses, the presence/absence of glasses does appear to be a latent variable. However, this suffers in comparison to the cVAE in two ways: (1) the presence/absence of glasses factor is entangled with the dominant variation in the datasets (related to skin color and background behind the face), making it difficult to adjust in isolation. (2) the latent feature does not relate to variation \textit{within} the population of glasses-wearing celebrities, but rather serves to differentiate between those who wear dark glasses and those who wear light/no glasses.

\begin{figure}[H]
\centering
\includegraphics[width=0.39\linewidth]{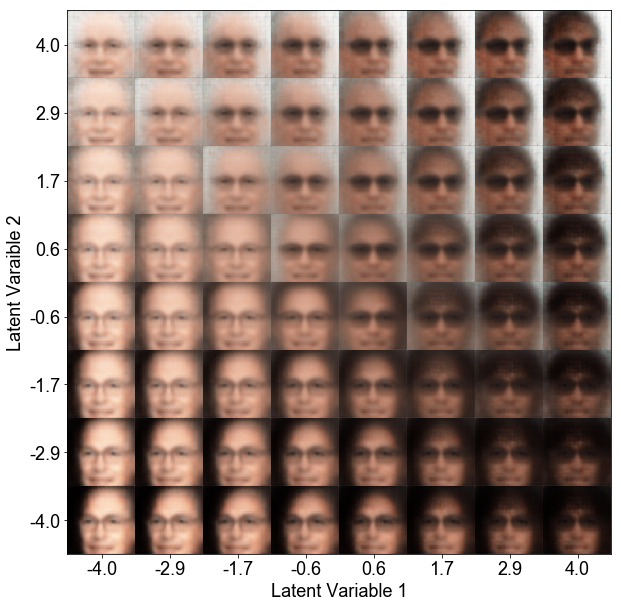}
\end{figure}

Similarly, in Fig. \ref{fig:celeb}(c), we showed that we could isolate latent variables related to eyeglasses by using a contrastive VAE. For comparison, here we show the reconstructions of a standard VAE trained \textit{only} on the target dataset of celebrities with hats. Because we are restricting ourselves to celebrities with hats, the presence/absence of hats does appear to be a latent variable. However, this suffers in comparison to the cVAE in two ways: (1) the presence/absence of hats factor is entangled with the dominant variation in the datasets (related to skin color and background behind the face), making it difficult to adjust in isolation. (2) the latent feature does not relate to variation \textit{within} the population of hat-wearing celebrities, but rather serves to differentiate between those who wear white hats and those who wear dark/no hats. 

\begin{figure}[H]
\centering
\includegraphics[width=0.39\linewidth]{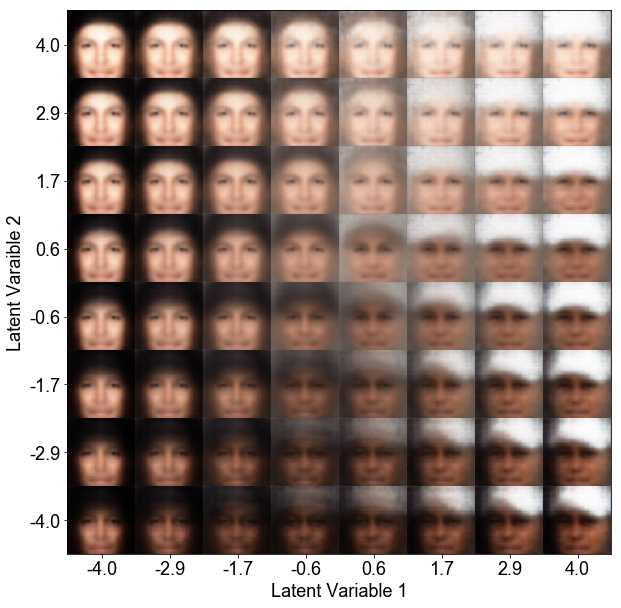}
\end{figure}

\clearpage
\section{Derivation of Variational Lowerbound}
\label{appendix:derivation}

In this appendix, we derive the variational lower bounds presented in (\ref{eq:target_likelihood}) and (\ref{eq:background_likelihood}) in the main text. We start by writing the log-likelihood for each sample in the target as:
\begin{align}
    \log{p_{\theta}(\xb^{(i)})} &= \log \sum_{\sbb, \zb} {p_{\theta}(\xb^{(i)}, \sbb, \zb)} \\
    &= \log \sum_{\sbb, \zb} \frac{p_{\theta}(\xb^{(i)}, \sbb, \zb) \cdot  q_{\phi}(\xb^{(i)} \vert \sbb, \zb)}{q_{\phi}(\xb^{(i)} \vert \sbb, \zb)} \\
    &\ge  \sum_{\sbb, \zb}   q_{\phi}(\xb^{(i)} \vert \sbb, \zb) \log \frac{p_{\theta}(\xb^{(i)}, \sbb, \zb) }{q_{\phi}(\xb^{(i)} \vert \sbb, \zb)}\label{eq:line8}\\
    &=  \sum_{\sbb, \zb}   q_{\phi}(\xb^{(i)} \vert \sbb, \zb) \log \frac{p_{\theta}(\xb^{(i)} \vert \sbb, \zb) \cdot p(\sbb, \zb) }{q_{\phi}(\xb^{(i)} \vert \sbb, \zb)}\\
    &=  \sum_{\sbb, \zb}   q_{\phi}(\xb^{(i)} \vert \sbb, \zb) \log p_{\theta}(\xb^{(i)}  \vert \sbb, \zb) + q_{\phi}(\xb^{(i)} \vert \sbb, \zb) \log \frac{p(\sbb, \zb) }{q_{\phi}(\xb^{(i)}, \sbb, \zb)}\\
    &=  \E_{q_{\phi}(s, z)}[\log p_\theta(\xb^{(i)} | \sbb, \zb)] -  \KL(q_{\phi}(\sbb, \zb | \xb^{(i)}) \Vert p(\sbb, \zb))\\    
    &=  \E_{q_{\phi_s}(s) q_{\phi_z}(z)}[\log p_\theta(\xb^{(i)} | \sbb, \zb)] -  \KL(q_{\phi_s}(\sbb | \xb^{(i)}) \Vert p(\sbb)) - \KL(q_{\phi_z}(\zb | \xb^{(i)}) \Vert p(\zb))\label{eq:line12}
\end{align}
where, in (\ref{eq:line8}), we have used Jensen's inequality, and in (\ref{eq:line12}), we have used the independence of $\zb$ and $\sbb$ to decompose the joint KL divergence into the sum of two marginal KL divergence. 


This gives us the variational bound in (\ref{eq:target_likelihood}). The derivation for the log-likelihood of the background data, (\ref{eq:background_likelihood}), is even simpler, as it consists of only one set of latent variables; it follows the same derivation as the standard variational autoencoder \citep{kingma2013auto}. 

\clearpage
\section{Total Correlation Loss}
\label{appendix:total_correlation}

In this appendix, we describe the total correlation loss in more detail, and how we compute it efficiently with contrastive VAEs. The total correlation loss is defined as:
\begin{align}
\text{TC} = - \KL(\bar{q} \Vert q_{\phi_s}(\sbb | \xb_i) \cdot q_{\phi_z}(\zb | \xb_i)),
\end{align}
where $\bar{q} \eqdef q_{\phi_s, \phi_z}(\sbb, \zb | \xb_i)$. This quantity cannot be easily approximated during stochastic gradient descent, since its batch estimate suffers from computational issues (see \citet{kim2018disentangling}, Appendix D for more details). Instead, we use the density-ratio trick to estimate the total correlation.

The key idea behind the density-ratio trick \citep{nguyen2010estimating} is to train a classifier that tries to predict whether a sample $[\sbb, \zb]$ is drawn from $\bar{q}$ or from $q_{\phi_s}(\sbb | \xb_i) \cdot q_{\phi_z}(\zb | \xb_i)$. We simulate samples from the latter term by randomly shuffling the $\zb$ and $\sbb$ dimensions.

The probabilities outputted by such a classifier can be used to estimate the desired KL term. Therefore, in our algorithm, we train a simple logistic regression classifier $D_\psi$ along with the cVAE. This classifier is included in the pseudocode shown in Fig. \ref{fig:method}(b). We compute the following quantity as the estimate of the total correlation:
\begin{align}
\text{TC} \approx \sum_{i} \log{( D_{\psi}(\vb^{(i)}) / 1-D_{\psi}(\vb^{(i)}))},
\end{align}
where $\vb^{(i)}$ is the concatenation of $[\sbb^{(i)}, \zb^{(i)}]$, and the summation is carried over the batch of samples.
\clearpage
\section{Additional Quantitative Results on the Grassy-MNIST Dataset}
\label{appendix:high_dimension_vae}

In this appendix, we include two sets of results. On the left, we train a standard VAE in four dimensions on the target dataset in Grassy-MNIST. We observe that even with four latent dimensions (the same total number of dimensions as the contrastive VAE), we still do not observe any latent dimensions which individually allow any kind of discrimination of the class digit. The top left plot shows the first and second latent dimensions, while the bottom left plot shows the third and fourth latent dimensions. 

On the right, top, we show the same result as in Fig. \ref{fig:mnist-quant}(a, top), in which we show the salient latent variables learned by the cVAE can be used to discriminate between digits. Additionally, in the bottom right, we also show the irrelevant variables learned by the cVAE. As expected, these do not vary in significant way among the classes of digits.

\begin{figure}[H]
\centering
\subfloat{\includegraphics[width=0.89\linewidth]{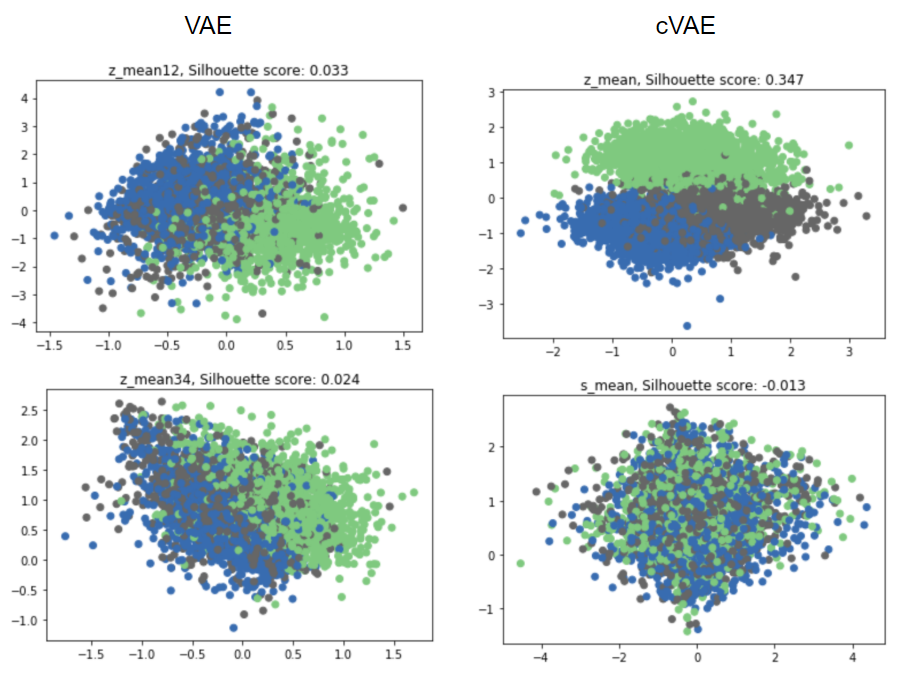}} 

\end{figure}

\clearpage
\section{Architecture Details}
\label{appendix:architecture}

For the experiments in this paper, we use two kinds of architectures for the VAEs/cVAEs. For the experiments with the Grassy-MNIST dataset, as well as the RNA-Seq dataset, we use a fully-connected architecture. The second kind of architecture includes has convolutional layers in the encoder and transpose-convolutional layers in the decoder. This architecture is used for experiments with the the CelebA dataset. The layers in each architecture are written below.

\textbf{Layers in the fully-connected VAE/cVAE:}
\begin{itemize}
    \item input shape of 784 [for \textbf{Grassy-MNIST}] and 500 [for \textbf{RNA-Seq}]
    \item intermediate layer of dimensionality 128
    \item (salient) latent feature layer with dimensionality 2 (in the cVAE, there are irrelevant feature layers whose dimensionality varies from 1 to 6 depending on the experiment)
    \item intermediate layer of dimensionality 128
    \item output layer of the same dimensionality as the input layer
\end{itemize}

\textbf{Layers in the convolutional VAE/cVAE:}
\begin{itemize}
    \item input shape of (64, 64, 3)
    \item 2D-convolutional layer with kernel size (5, 5) and a stride of (2, 2) with 32 channels
    \item 2D-convolutional layer with kernel size (5, 5) and a stride of (2, 2) with 64 channels
    \item intermediate fully-connected layer of dimensionality 128
    \item salient latent feature layer with dimensionality  2 (in the cVAE, there are irrelevant feature layers with dimensionality 4)
    \item intermediate fully-connected layer of dimensionality 128
    \item 2D-convolutional-transpose layer with kernel size (5, 5) and a stride of (2, 2) with 64 channels
    \item 2D-convolutional-transpose layer with kernel size (5, 5) and a stride of (2, 2) with 64 channels
    \item final output of of (64, 64, 3)
\end{itemize}

\clearpage
\section{Complete Results for Fig. 5}
\label{appendix:fig_5}

In Fig. \ref{fig:rna}(a), the plots we show omit a few points for ease of visualizations. Here, we show the complete plots.

\begin{figure}[H]
\centering
\subfloat[]{\includegraphics[width=0.49\linewidth]{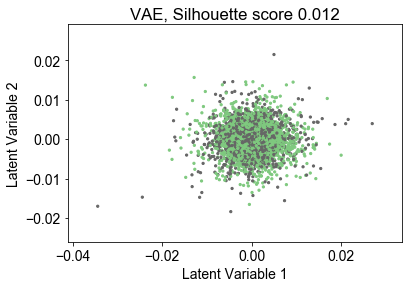}} 
\subfloat[]{\includegraphics[width=0.49\linewidth]{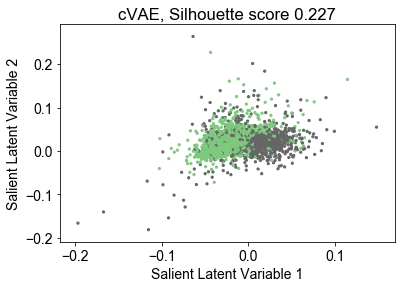}} 
\end{figure}

\clearpage
\section{Sensitivity Analyses with Respect to Dataset Composition}
\label{appendix:sensitivity}

We conducted two additional experiments in order to characterize the sensitity of the contrastive VAE to the \textit{composition} of the target and background datasets. In the first experiment, we constructed a target dataset that included varying amounts of \textit{background} images (of only grass). The background datset was unchanged. We carried out the rest of the experiment as outlined in Section \ref{subsection:sensitivity}, in which we measured the resulting silhouette scores. As shown on the left, we found that that the contrastive VAE performance degraded as more background images were introduced in the target dataset -- which is not surprising, as the contrast between the target and background was weakened by the addition of background images to the target dataset.

In the second experiment, we did not change the target dataset, but instead changed the composition of the background dataset to include images from the target dataset. We measured the resulting silhouette scores. We found that that the contrastive VAE performance surprisingly did not seem to degrade as more target images were introduced in the background dataset, as shown on the right.

\begin{figure}[H]
\centering
\subfloat{\includegraphics[width=0.49\linewidth]{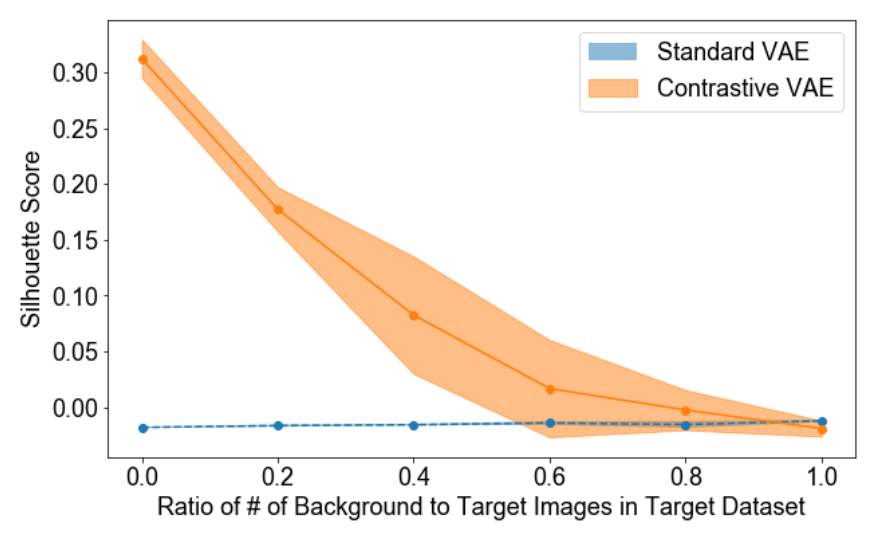}} 
\subfloat{\includegraphics[width=0.49\linewidth]{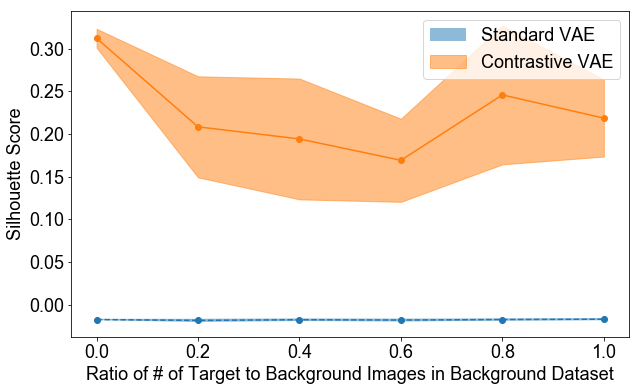}} 
\end{figure}

\end{onecolumn}

\end{document}